\documentclass[12pt,paper]{article}

\usepackage{cvpr}
\usepackage{times}
\usepackage{epsfig}
\usepackage{graphicx}
\usepackage{amsmath}
\usepackage{amssymb}
\usepackage{gensymb}
\usepackage{multirow}
\usepackage{subfig}
\usepackage{algorithm2e}

\cvprfinalcopy 

\ifcvprfinal\fi

\begin{document}

\title{Descriptor learning for omnidirectional image matching}

\author{Jonathan Masci$^{1,2,3}$ \\
{\tt \small jonathan@idsia.ch} \\
\and Davide Migliore$^{1,4}$ \\
{\tt \small davide.migliore@gmail.com}
\and Michael M. Bronstein$^{2}$ \\
{\tt\small michael.bronstein@usi.ch}
\and J{\"u}rgen Schmidhuber$^{1,2,3}$\\
{\tt \small juergen@idsia.ch} \\
\vspace{3mm}\\
\small $^1$Istituto Dalle Molle di Studi sull'Intelligenza Artificiale (IDSIA), Manno, Switzerland\\
\small $^2$Faculty of Informatics, Universit{\`a} della Svizzera Italiana (USI), Lugano, Switzerland\\
\small $^3$Scuola Universitaria Professionale della Svizzera Italiana (SUPSI), Lugano, Switzerland\\
\small $^4$ Evidence Srl, Pisa, Italy
}


\maketitle

\thispagestyle{empty} 

\begin{abstract}

Feature matching in omnidirectional vision systems is a challenging problem, mainly because complicated optical systems make the theoretical modelling of invariance and construction of invariant feature descriptors hard or even impossible. 
In this paper, we propose learning invariant descriptors using a training set of similar and dissimilar descriptor pairs. We use the similarity-preserving hashing framework, in which we are trying to map the descriptor data to the Hamming space preserving the descriptor similarity on the training set. A neural network is used to solve the underlying optimization problem. Our approach outperforms not only straightforward descriptor matching, but also state-of-the-art similarity-preserving hashing methods.

\end{abstract}

\section{Introduction}

Feature-based matching between images has become a standard approach in computer vision literature in the last decade, in many respects due to the introduction of stable and invariant feature detection and description algorithms such as SIFT~\cite{Lowe04g} and
similar  methods~\cite{Miko05,Bay08,Tola10}.  
The usual assumption guiding the design of feature descriptors is invariance across viewpoints, which should guarantee that the same feature appearing in two different views has the same descriptor. 
Since perspective transformations are approximately locally affine, it is common to construct affine-invariant descriptors \cite{Cuiping}.

While being a good model in many cases, affine invariance is not sufficiently accurate in cases of wide baseline (very different view points) or even more complicated setting of optical imperfections such as lens distortions, blur, etc. In particular, in omnidirectional vision systems the distortion is introduced intentionally (e.g., using a parabolic mirror \cite{mei07a}) to allow a $360^{\circ}$ view. 
Designing invariant descriptors for such cases is challenging, as the invariance is complicated and cannot be easily modeled.

An alternative to `invariance-by-construction' approaches which rely on a simplified invariance model is to {\em learn} the descriptor invariance from examples. Recent work of Strecha {\em et al.} \cite{ldahash} showed very convincingly that such approaches can significantly improve the performance of existing descriptors.

In this paper, we consider the learning of invariant descriptors for omnidirectional image matching. We construct a training set of similar and dissimilar descriptor pairs including strong optical distortions, and use a neural network to learn a mapping from the descriptor space to the Hamming space preserving similarity on the training set. Experimental results show that our approach outperforms not only straightforward descriptors, but also other similarity-preserving hashing methods. The latter observation is explained by the suboptimality of existing approaches which solve a simplified optimization problem.

The main contribution of this paper is two-fold. First, we formulate a new similarity-sensitive hashing algorithm. Second, we use this approach to learn smaller invariant descriptors suitable for feature matching in omnidirectional images. 
The rest of the paper is organized as follows. In Section~2, we overview the related works. Section~3 is dedicated to metric learning and similarity-preserving hashing methods. In Section~4, we describe our NNhash approach. Section~5 contains experimental results. Finally, Section~6 discusses potential future work and concludes the paper.

\section{Background}

Although feature-based correspondence problems have been investigated in depth for standard perspective cameras, 
omnidirectional image matching still remains an open problem, largely because of the complicated geometry introduced by lenses and curved mirrors. Broadly speaking, the existing approaches either try to reduce the problem to the simpler perspective setting, or design special descriptors suitable for omnidirectional images. 

Svoboda {\em et al.} \cite{Svoboda2001} proposed to use adaptive windows around interest points to generate normalized patches with the assumption that the displacement of the omnidirectional system is smaller than the depth of the surrounding scene.
Nayar \cite{Nayar1997} showed that, given the mirror parameters, it is possible to generate a perspective version of the omnidirectional image and Mauthner {\em et al.} \cite{Mauthner2006} used this approach to generate perspective representation of each interest point region.
This unwarping procedure removes the non-linear distortions and enables the use of algorithms designed for perspective cameras.
Micusik and Pajdla \cite{Micusik2006} 
checked the candidate correspondences between two views using the RANSAC algorithm and the epipolar constraint \cite{Geyer2007}.
%
%
Construction of scale-space by means of diffusion on manifolds was used in \cite{BogdanovaVandergheynst2007,Hansen2010,Cruz2009} for the construction of local descriptors. 
%
Puig {\em et al.} \cite{Puig2011} integrated the sphere camera model with the partial differential equations on manifolds framework. 

Another possible solution is to consider different kind of features to exploit particular invariance in omnidirectional systems, for example,  extracting one-dimensional features \cite{Briggs2006} or vertical lines \cite{Scaramuzza2009} and defining descriptors suitable for omnidirectional images.

More recently, it was shown in \cite{ldahash} that one can approach the design of invariant descriptors from the perspective of {\em metric learning}, constructing a distance between the descriptor vectors from a training set of similar  and dissimilar  pairs \cite{athitsos04,wang10}. 
In particular, {\em similarity-preserving hashing} methods \cite{Indyk99,Shakhnarovich05,weiss2009spectral,Kulis09,raginsky-locality} 
were found especially attractive for descriptor learning, as they significantly reduce descriptor storage and comparison complexity.
These  methods have  also  been applied  to image search \cite{jain08_,Torralba08a,jegou2009packing,jegou2008hamming,jegou2010product,wang10a}, video copy detection \cite{vnome}, and shape retrieval \cite{ovsjanikov2009shape}.

In \cite{SalHinSH07}, binary codes were produced using a restricted Boltzmann machine and in \cite{weiss2009spectral} using spectral hashing in an unsupervised setting. The authors showed that the learnt binary vectors capture the similarities of the data. With such an approach it is however impossible to explicitly provide information about data similarities. Since in our problem it is easy to produce labeled data, supervised metric learning is advantageous. 


\newcommand{\bm}[1]{\boldsymbol{#1}}
\newcommand{\bb}[1]{\boldsymbol{\mathrm{#1}}}

\newcommand{\pp}{\mathbf{p}}
\newcommand{\xx}{\mathbf{x}}
\newcommand{\yy}{\mathbf{y}}
\newcommand{\PP}{\mathbf{P}}
\newcommand{\QQ}{\mathbf{Q}}
\newcommand{\Dd}{\mathbf{D}}
\newcommand{\Rr}{\mathbf{R}}
\newcommand{\Ii}{\mathbf{I}}

\newcommand{\SSS}{\mathbf{S}}
\newcommand{\ttt}{\mathbf{t}}

\newcommand{\EE}{\mathbb{E}}
\newcommand{\RR}{\mathbb{R}}
\newcommand{\SSSS}{\mathbb{S}}
\newcommand{\ZZ}{\mathbb{Z}}
\newcommand{\HH}{\mathbb{H}}

\newcommand{\uu}{\bm{\mathrm{u}}}
\newcommand{\ww}{\bm{\mathrm{w}}}
\newcommand{\qq}{\bm{\mathrm{q}}}
\newcommand{\hh}{\bm{\mathrm{h}}}

\newcommand{\Ssigma}{\bm{\Sigma}}

\newcommand{\ssigma}{\bm{\sigma}}
\newcommand{\rrho}{\bm{\rho}}
\newcommand{\llambda}{\bm{\lambda}}
\newcommand{\mmu}{\bm{\mu}}
\newcommand{\xxi}{\bm{\xi}}
\newcommand{\aalpha}{\bm{\alpha}}

\newcommand{\ones}{\bm{\mathrm{1}}}

\newcommand{\Tr}{\mathrm{T}}

\section{Similarity preserving hashing}

Given a set of keypoint descriptors, represented as $n$-dimensional vectors in  $\RR^n$, the problem of {\em metric learning} is to find their representation in some metric space
$(\ZZ,d_\ZZ)$ by means of a map of the form $\yy : \RR^n \rightarrow (\ZZ,d_\ZZ)$.
The metric $d_\ZZ \circ (\yy \times \yy)$ parametrizes the similarity between the feature descriptors, which may be difficult to compute in the original representation.
%
%
%
Typically, $(\ZZ,d_\ZZ)$ is fixed and $\yy$ is the map we are trying to find in such a way that, 
given a set $\mathcal{P}$ of pairs of descriptors from corresponding points in different images 
(\emph{positives}) and a set $\mathcal{N}$ of pairs of descriptors from different points (\emph{negatives}), we have 
$d_\ZZ(\yy(\xx),\yy(\xx^+)) \approx 0$ for all $(\xx,\xx^+)\in \mathcal{P}$ and $d_\ZZ(\yy(\xx),\yy(\xx^-)) \gg 0$ for all $(\xx,\xx^-)\in \mathcal{N}$ with high probability.

A particular setting of this problem, where $\ZZ = \{\pm 1\}^m$ is the $m$-dimensional space of binary strings and $d_{\mathbb{H}^m}(\yy,\yy') = \frac{m}{2} - \frac{1}{2} \sum_{i=1}^m \mathrm{sign}(y_i y'_i)$ is the Hamming metric, the problem is referred to as {\em similarity-preserving hashing}.  
%
Here, we limit our attention to affine embeddings of the form
\begin{eqnarray}
\yy= \mathrm{sign}(\PP \xx + \ttt) \ ,
\label{eq:model}
\end{eqnarray}
where $\PP$ is an $m\times n$ matrix and $\ttt$ is an $m \times 1$ vector. 
Our goal is to find such $\PP$ and $\ttt$ that minimize one of the following cost functions,  
\begin{eqnarray}
L_\mathrm{c}(\PP,\ttt) &=& \EE \{ \yy(\xx)^\Tr \yy(\xx^-) - \alpha \yy(\xx)^\Tr \yy(\xx^+) \},\,\,\mathrm{or} \nonumber\\
L_\mathrm{d}(\PP,\ttt) &=& \EE \{ \alpha \|\yy(\xx) -  \yy(\xx^+)\|^2 - \| \yy(\xx) - \yy(\xx^-)\|^2 \}\nonumber
\label{eq:cost_corr}
\end{eqnarray}
for $(\xx,\xx^+) \in \mathcal{P}$ and $(\xx,\xx^-) \in \mathcal{N}$. 
Both cost functions try to map positives as close as possible to each other (expressed as large correlations or small distance), and negatives as far as possible from each other (small correlation or large distance), in order to ensure low false positive (FPR) and false negative (FNR) rates. $\alpha>0$ is a parameter determining the tradeoff between the FPR and FNR. 
In practice, the expectations are approximated as means on some sufficiently large training set.

The problem $\min_{\PP,\ttt}L(\PP,\ttt)$ is a non-linear non-convex optimization problem without an obvious simple solution.
It is commonly approached by the following two-stage relaxation: first, approximate the map $\yy \approx \PP\xx$ by removing the sign and the offset vectors, minimizing 
\begin{eqnarray}
\hat{L}_\mathrm{c}(\PP) &=& \EE \{ (\PP\xx)^\Tr (\PP\xx^-) - \alpha (\PP\xx)^\Tr (\PP\xx^+) \},\,\,\mathrm{or} \nonumber\\
\hat{L}_\mathrm{d}(\PP) &=& \EE \{ \alpha \|\PP(\xx - \xx^+)\|^2 - \| \PP(\xx - \xx^-)\|^2 \}\nonumber
\label{eq:cost_corr_}
\end{eqnarray}
w.r.t. to $\PP$ (introducing some regularization, e.g., $\bb{P}^\Tr \bb{P} = \bb{I}$, in order to avoid a trivial solution $\bb{P} = 0$). Second, fix $\PP^* = \arg\min_{\PP}\hat{L}(\PP)$ and solve $\ttt^* = \arg\min_{\ttt}L(\PP^*,\ttt)$ w.r.t. $\ttt$. 
To further simplify the problem, it is also common to assume {\em separability}, thus solving independently for each dimension of the hash. 


\subsection{Similarity-sensitive hashing (SSH)}

In \cite{Shakhnarovich05}, the above strategy was used for the approximate minimization of the cost $L_\mathrm{c}$.
The computation of optimal parameters $\PP$ and $\ttt$ was posed as a boosted binary classification problem,
where $d_\mathbb{H}(\yy,\yy')$ acts as a strong binary classifier, and each dimension of the linear projection
$\mathrm{sign}(\pp_i \xx + t_i)$ is considered a weak classifier (here, $\pp_i$ denotes the $i$th row of $\PP$). This way, AdaBoost can be used to find a greedy approximation of the minimizer
of $L_\mathrm{c}$ by progressively constructing $\PP$ and $\ttt$.
At the $i$-th iteration, the $i$-th row of the matrix $\PP$ and the $i$-th element of the vector $\ttt$ are found minimizing a weighted version of $L_\mathrm{c}$.
Since the problem is non-linear, such an optimization is a challenging problem. In \cite{Shakhnarovich05}, random projection directions were used.
A better method for projection selection similar to linear discriminative analysis (LDA) was proposed \cite{vnome,paragios}. 
Weights of false positive and false negative pairs are increased, and weights of true positive and true negative pairs are decreased, using the standard AdaBoost reweighting scheme \cite{Freund95}. 

\subsection{Covariance difference hashing (diff-hash)}

In \cite{ldahash}, it was observed that the minimization $\min_{\PP}\hat{L}_\mathrm{c}(\PP)$ can be written as
\begin{eqnarray}
\min_{\bb{P}} \mathrm{tr}\{\bb{P}( \alpha \bb{C}_+  - \bb{C}_- ) \bb{P}^\Tr \} \,\,\, \mathrm{s.t.} \,\,\, \bb{P}^\Tr \bb{P} = \bb{I},
\end{eqnarray}
where $\bb{C}_\pm = \mathbb{E}\{ (\bb{x}-\bb{x}^\pm) (\bb{x}-\bb{x}^\pm)^\Tr \}$ are the covariance matrices of the differences of the positive and negative pairs of vectors.
Requiring an orthonormal projection matrix $\bb{P}$, the problem has a closed-form solution consisting of the $m$ smallest eigenvectors of $( \alpha \bb{C}_+  - \bb{C}_- )$, and is thus also a separable problem. 
Once the projection is found in this way, the threshold vector $\bb{t}$ maximizing the sum of the false positive and false negative rates is selected. This second stage also turns out separable in each dimension.
In \cite{kdiffhash}, a more generic kernelized version of diff-hash (kdiff-hash) was shown.  

\subsection{LDAHash}

A similar method was derived in \cite{ldahash} by transforming the coordinates as  $\bb{C}_-^{-1/2}\xx$, which allows to write $\min_{\PP}\hat{L}_\mathrm{d}(\PP)$ as
\begin{eqnarray}
\min_{\bb{P}} \mathrm{tr}\{\bb{P}( \bb{C}_+  \bb{C}^{-1}_- ) \bb{P}^\Tr \} \,\,\, \mathrm{s.t.} \,\,\, \bb{P}^\Tr \bb{P} = \bb{I}. 
\end{eqnarray}
This approach resembles linear discriminant analysis (LDA), hence the name LDAhash.
Requiring an orthonormal projection matrix $\bb{P}$, the problem has a separable closed-form solution consisting of the $m$ smallest eigenvectors of $( \bb{C}_+  \bb{C}^{-1}_- )$.

\section{Neural network hashing (NNhash)}

The problem of existing and most successful similarity-preserving hashing approaches such as LDA- or diff-hash is that they do not solve the optimization problem $\min_{\PP,\ttt}L(\PP,\ttt)$ but rather its relaxation. As a result, the parameters $\PP^*, \ttt^*$ found by these methods in the aforementioned two-stage separable scheme is suboptimal, i.e., $L(\PP^*,\ttt^*) > \min L$. Our experience shows that in some cases, the suboptimality is dramatic (at least an order of magnitude). 


A way of solving the `true' optimization problem is by formulating it in the neural network (NN) framework and exploiting numerous optimization techniques and heuristics developed in this field. 
%
Since we have a way of cheaply producing labeled data, we decide to adopt the {\em siamese network} architecture \cite{SchmidhuberPrelinger:93,hadsell-chopra-lecun-06} which, contrary to conventional models, receives two input patterns and minimize a loss function similar to equation~(\ref{eq:cost_corr_}),
\begin{eqnarray}
L_\mathrm{nn}(\PP,\ttt) &=& \frac{1}{2} \|\yy(\xx) -  \yy(\xx^+)\|^2 
				  + \frac{1}{2} (\max\{0, m - \| \yy(\xx) - \yy(\xx^-)\|\})^{2},
				 \label{eq:nncost}
\end{eqnarray}
where the constant $m$ represents the margin between dissimilar pairs. The margin is introduced as regularization to avoid the system from minimizing the loss just pulling two vector as far apart as possible.
The embedding is then learned to make positive pairs as close as possible and negative pairs at least at distance $m$. 

Network architecture of this type can be traced back to the work of Schmidhuber and Prelinger  \cite{SchmidhuberPrelinger:93} on problems of predictable classification. 
In \cite{hadsell-chopra-lecun-06}, siamese networks were used to learn an invariant mapping of tiny images directly from pixel representation, whereas in \cite{Taylor11} a similar approach is used to learn a model that is highly effective at matching people in similar pose which exhibits invariance to identity, clothing, background, lighting, shift and scale.
An advantage of such architecture is that one can create arbitrarily complex embeddings by simply stacking many layers in the network.
In all our experiments, in order to make a fair comparison to other hashing methods, we adopt a simple single layer architecture, wherein $\yy(\xx) = \mathrm{sign}(\PP\xx + \ttt)$. 
Network training attempts to find $\PP, \ttt$ that minimize $L_\mathrm{nn}$ (which is a regularized version of $L_\mathrm{d}$). Since we solve a non-linear problem without introducing any simplification or relaxation, the results are expected to be better compared to hashing methods described in Section~3. In the following, we refer to our method as {\em NNhash}. 

Since a binary output is required, we adopt $\mathrm{tanh}(\beta t) \approx \mathrm{sign}(t)$ as the non-linear activation function for our siamese network, which enforces binary vectors when either $m$ or the steepness $\beta$ of the function is increased.
Since the problem is highly non-convex, it is liable to local convergence, and thus there is no theoretical guarantee to find the global minimum. However, by initializing $\PP, \ttt$ by the solution obtained by one of the standard hashing methods, we have a good initial point that can be improved by network optimization, 
%

\section{Results}

\begin{figure*}[ht]
\begin{center}
\includegraphics[width=\linewidth]{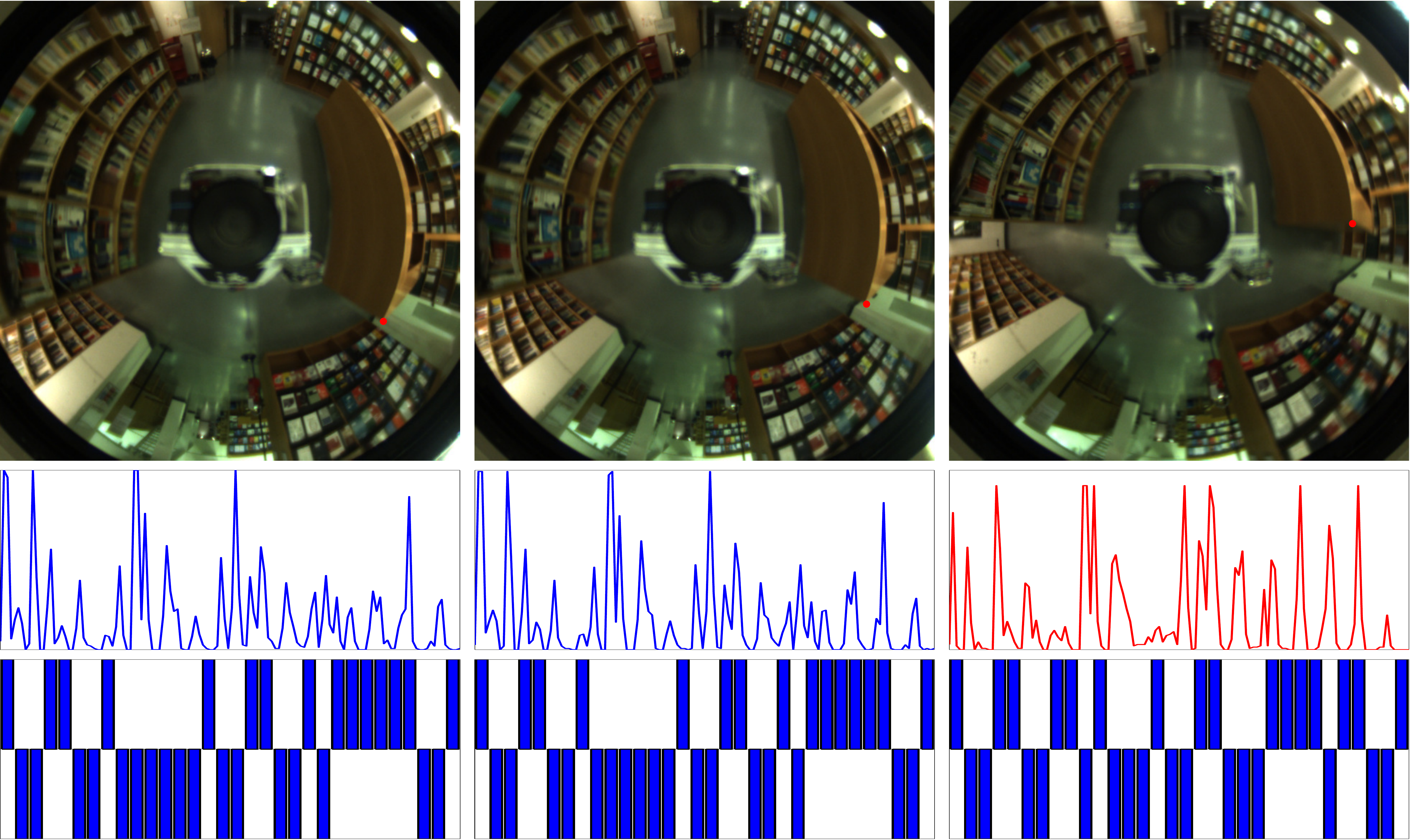}
\end{center}
\caption{A few frames from the Rawseeds dataset examplifying how a descriptor changes over time due to camera motion throughout the scene. First row: omnidirectional images of the indoor dataset, shown at times 1 (left), 5 (middle) and 50 (right). Second row: SIFT descriptors at point indicated in red. Third row: binary descriptors of length 32 produced by NNhash trained on outdoor images.}
\label{fig:educativeImage}
\end{figure*}

\subsection{Data} 
 
In our experiments, we used the Rawseeds dataset \cite{Bonarini_2006_IROS,Ceriani_2009_AutonomousRobots}. 
The dataset contained video sequences of a robot equipped with an omnidirectional camera system based on a parabolic mirror moving in an indoor and outdoor scene. The image undergoes significant distortion since different parts of the scene move from the central part of the mirror to the boundaries.

We used the toolbox of Vedaldi \cite{Vedaldi07} to compute SIFT features in each frame of the video. Since the robot movement is slow, the change between two adjacent frames in the dataset is infinitesimal, and SIFT features can be matched reliably. Tracking features for multiple frames, we  constructed the positive set as the transitive closure of these adjacent feature descriptor pairs. This way, the positive set included also descriptors distant in time, and, as a result of robot motion located at different regions in the image and thus subject to strong distortions. 
As negatives, we used features not belonging to the same track.

\begin{figure*}[!ht]
\begin{center}
\includegraphics[width=1\linewidth]{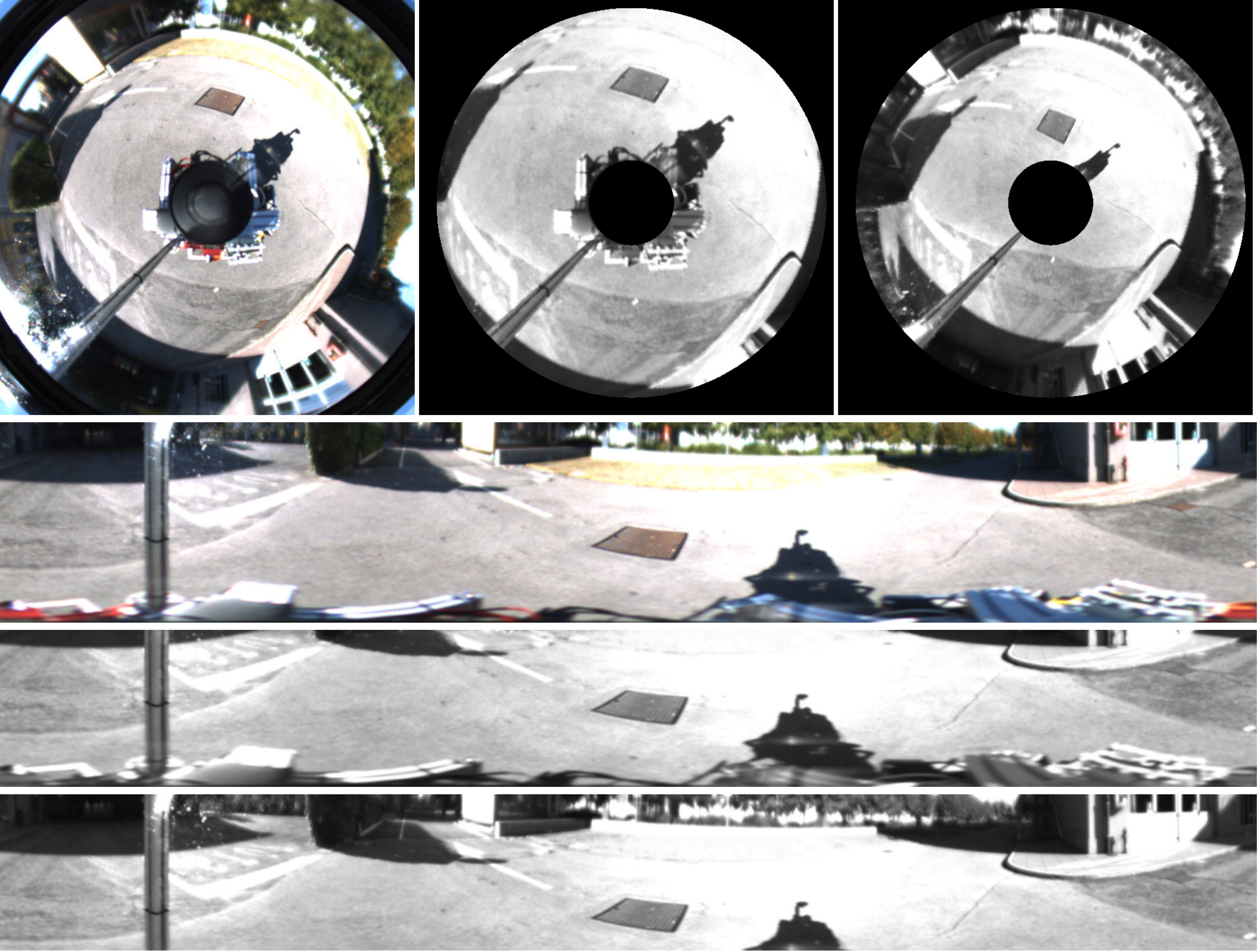}
\end{center}
\caption{Illustrative example of how synthetic data is generated. First row from left to right: the original omnidirectional image, the synthetic image from the first shift of 5 pixels, the synthetic image after 14 vertical shifts. Second to fourth rows: unwarped panorama images generated from images in the first row.}
\label{fig:synthEducative}
\end{figure*}

In addition to the Rawseeds dataset, we created synthetic omnidirectional datasets using panorama images that were warped simulating the effect of a parabolic mirror. The warping intentionally was not the same as in Rawseeds dataset. By moving the panorama image, we created synthetic motion with known pixel-wise groundtruth correspondence (Figure~\ref{fig:synthEducative}). The positive and negative sets for synthetic data were constructed as described above.

\subsection{Methods}

 We compared the SSH \cite{Shakhnarovich05}, diff-hash \cite{ldahash}, and our NNhash methods. 
%
For the NNhash training we used scaled conjugate gradient over the whole batch of descriptors, which we normalize in the range $[-1..1]$.
We used a margin $m = 5$ in all cases. The steepness factor for $\mathrm{tanh}$ is $\beta = 1$ in the case of $32$ bit while for $64$ bit we gradually increased it up to $3$ so to have a smooth binarization.
We reached convergence in about $50$ epochs in all cases.

\subsection{Performance degradation in time}

For this experiment, we constructed the training set using descriptors extracted from about $300$ consecutive frames of the outdoor sequence (similar results were obtained when using outdoor or synthetic data for training). 
We considered descriptors that could be tracked for at least $60$ consecutive frames and selected as positives pairs of descriptors belonging to these tracks. 
\begin{figure*}[!ht]
\begin{center}
\includegraphics[width=0.48\linewidth]{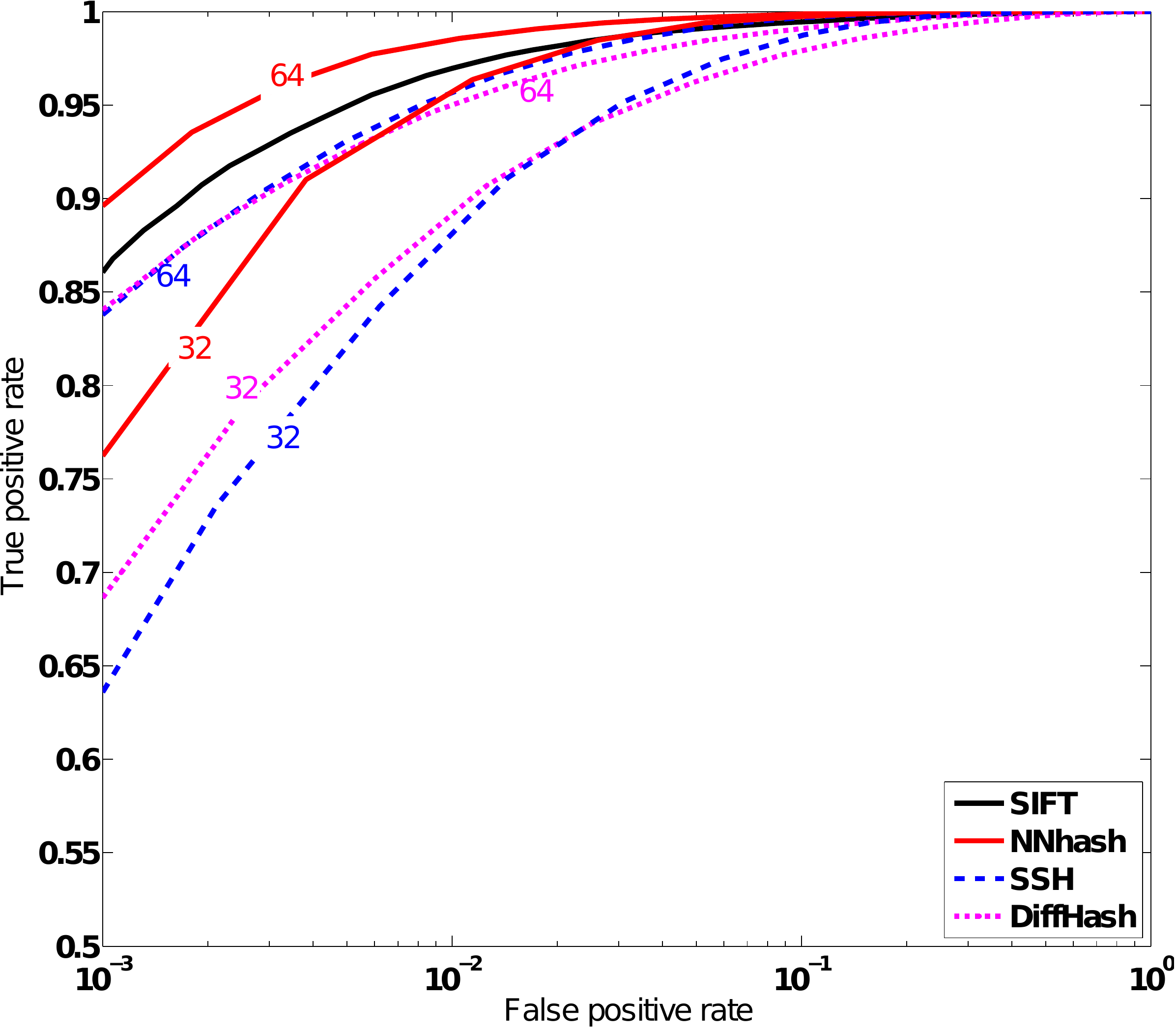} \hspace{2mm}
\includegraphics[width=0.48\linewidth]{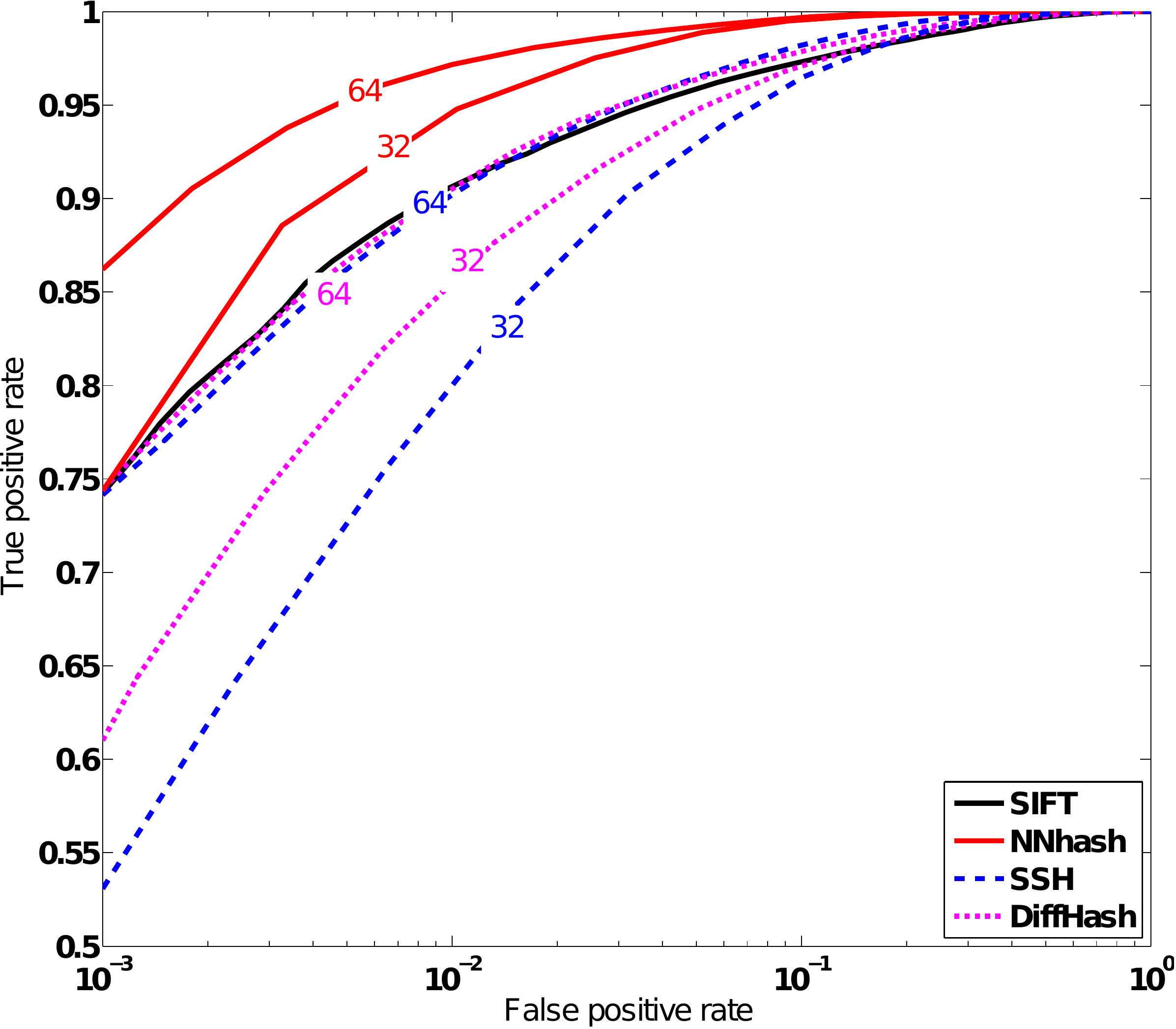}\\
\hspace{5mm}$10 \leq \Delta t \leq 30$ \hspace{66mm} $20 \leq \Delta t \leq 40$\\
\end{center}
\caption{ROC curve for the outdoor dataset, with frames taken at various distance $\Delta t$. Each hashing method is shown with $32$ and $64$ bits. 
Note significant performance degradation of SIFT and only minor performance degradation of NNhash. }
\label{fig:roc_out_out_10_30}
\end{figure*}

To avoid bias, we selected pairs of descriptors in frames $t_i, t_j$ in such a way that the time difference $\Delta t = |t_i - t_j|$ between the frames was uniformly distributed. 
%
The training was performed on a positive set of size $10^5$ and on a negative set of size $10^6$ to produce hashes of length $32$ and $64$ bits.

Testing was performed on a different portion of the same sequence, where frames at distance $10 \leq \Delta t \leq 30$ (Figure~\ref{fig:roc_out_out_10_30}, left) and $20 \leq \Delta t \leq 40$ (Figure~\ref{fig:roc_out_out_10_30}, right) were used. 
A few phenomena can be observed in Figure~\ref{fig:roc_out_out_10_30} showing the ROC curves of straightforward SIFT matching using the Euclidean distance and matching of learned binary descriptors using the Hamming distance. 
First, we can see that even with very compact descriptors (as small as $64$ bit, compared to $1024$ bit required to represent SIFT) 
we match or outperform SIFT.
These results are consistent with the study in  \cite{ldahash}. 
Second, we observe that NNhash significantly outperforms other hashing methods for the same number of bits. This is a clear indication that SSH and diff-hash methods are finding a suboptimal solution by solving a relaxed problem, while NNhash attempts to solve the full non-linear non-convex optimization problem. 
  
Comparing Figure~\ref{fig:roc_out_out_10_30} (left and right) and Tables~\ref{tab:fprfnr1030}--\ref{tab:fprfnr4060}, we can observe how the matching performance degrades if we increase the time between the frames (from $10-30$ frames to $20-40$ frames). Because of significant distortions caused by the parabolic mirror, objects moving around the scene appear differently. This phenomenon is especially noticeable when the distance between the frames ($\Delta t$) is large. SIFT shows significant degradation, while NNhash, trained on a dataset including positive pairs at distances up to $\Delta t = 60$ degrades only slightly (even a $32$-bit NNhash performs better than SIFT). 
This is a clear indication that we are able to learn feature invariance. 

Finally, Figure~\ref{fig:matches_cmp} shows a visual example of feature matching using different methods. NNhash produces matches most similar to the groundtruth (shown in green).

%
%
%

\begin{table}[htdp]
\begin{center}
\begin{tabular}{rccccc}
		& $m$ & \bf{EER}	 & \bf{ FPR@1\%} 	& \bf{ FPR@0.1\%}  \\
\hline
{\bf SIFT}	& 1024	&  1.91\% & 3.08\%    &  13.87\%    \\
\hline
{\bf NNhash}    & 32 &  {\bf 1.66\%}   & 3.77\%  &  23.81\% 	\\
		& 64 &  {\bf 1.31\%}	 & {\bf 1.92\%}	 & {\bf 9.48\%} 	\\
\hline
{\bf DiffHash} & 32 & 4.41\%	& 9.36\%	& 29.95\%   \\
		& 64 	& 2.57\%	& 5.17\%  	& 18.30\%   \\
\hline
{\bf SSH}	& 32 &  4.02\%   & 15.64\%	&  36.41\%   \\	
		& 64 &  2.22\%   & 4.90\%  &  16.74\%  \\ 
\hline
\end{tabular}
\end{center}
\vspace{-3mm}
\caption{\label{tab:fprfnr1030} Descriptor matching performance using different methods and descriptor size for frames with time range $10 \le \Delta t \le 30$.}
\end{table}

\begin{table}[htdp]
\begin{center}
\begin{tabular}{rccccc}
		& $m$ & \bf{EER}	 & \bf{ FPR@1\%} 	& \bf{ FPR@0.1\%}  \\
\hline
{\bf SIFT}	& 1024	&  3.31\% & 7.47\% & 27.94\%    \\
\hline
{\bf NNhash}    & 32 &  {\bf 2.70\%}   & {\bf 6.98\%}  &  {\bf 24.98\%}	\\
		& 64 &  {\bf 2.38\%}	 & {\bf 4.54\%}	& {\bf 14.22\%} 	\\
\hline
{\bf DiffHash} & 32 	& 5.17\%	& 12.55\%  & 37.49\%   \\
		& 64 &  3.69\%	 & 8.75\%  &  27.34\% 	 \\
\hline
{\bf SSH}	& 32	&   5.52\%  & 24.10\%  & 47.29\%   \\
		& 64 &  3.46\% 	 & 9.48\%  & 27.66\%  	 \\
\hline
\end{tabular}
\end{center}
\vspace{-3mm}
\caption{\label{tab:fprfnr4060} Descriptor matching performance using different methods and descriptor size for frames with time range $20 \le \Delta t \le 40$.}
\end{table}

\subsection{Generalization}
To test for generalization we perform experiments of transfer learning from outdoor data to indoor data and from synthetic data to real data. 
%

\begin{figure*}[!ht]
\begin{center}
\includegraphics[width=.45\linewidth]{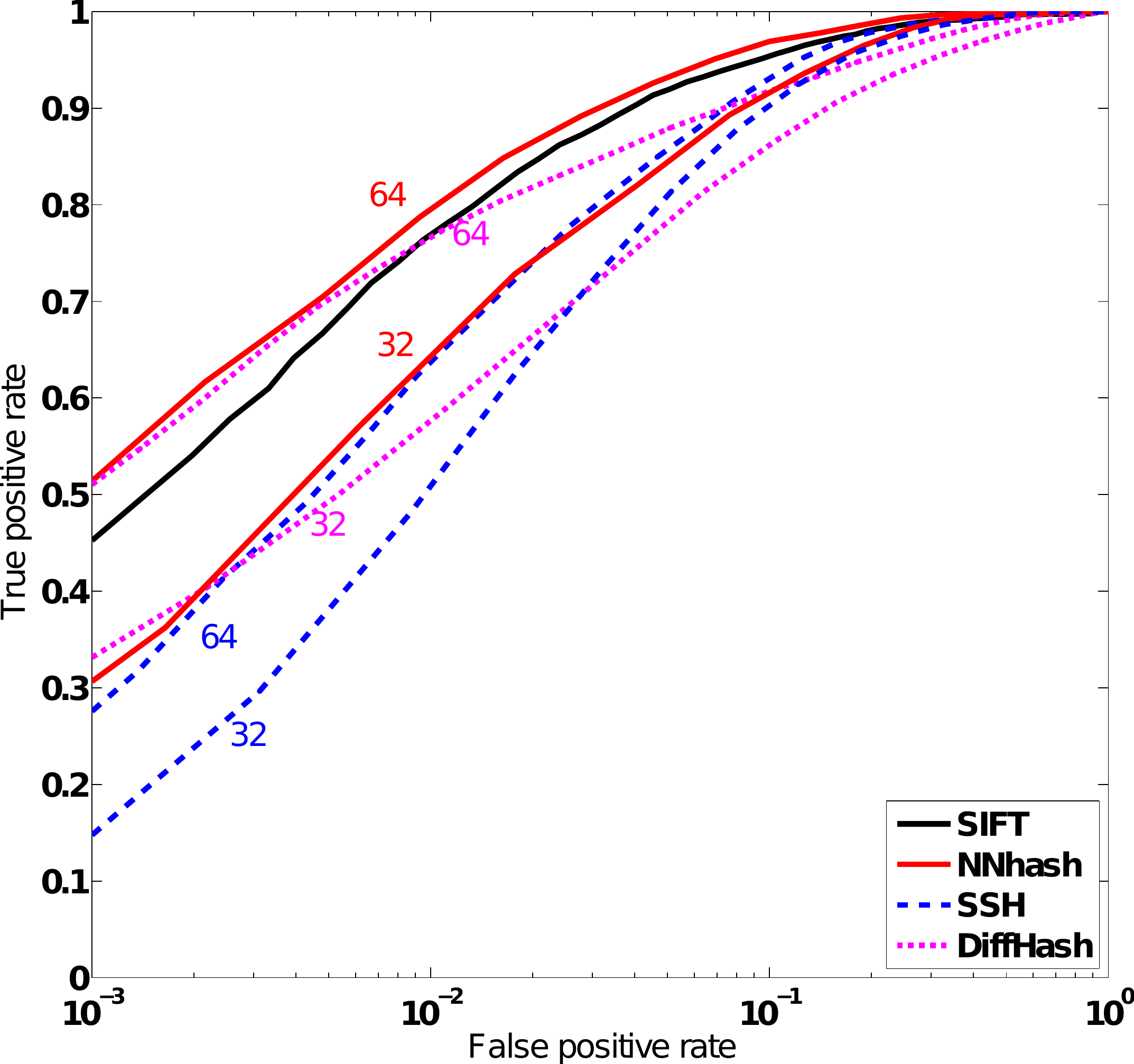}
\hspace{2mm}
\includegraphics[width=.48\linewidth]{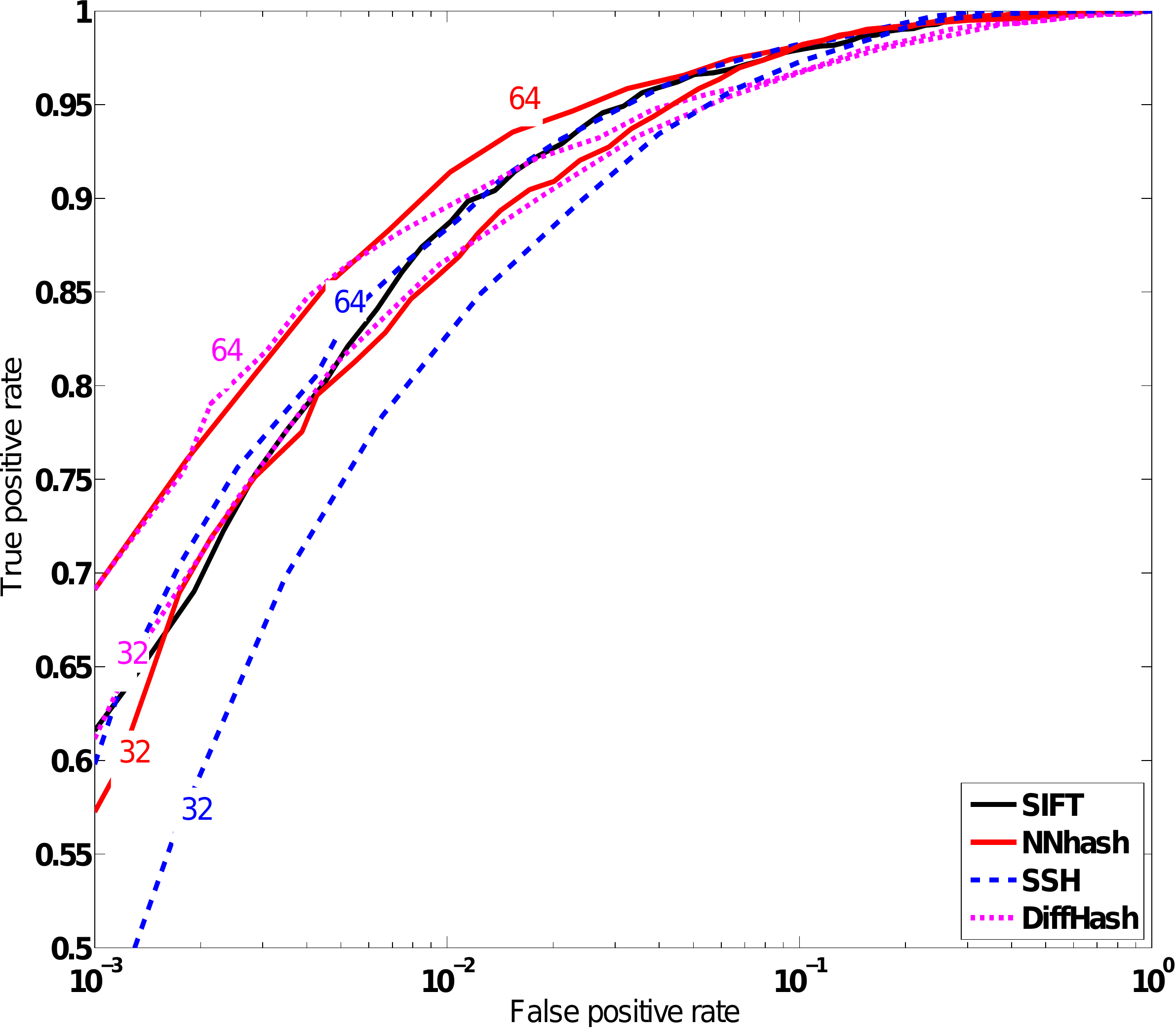}\\
\vspace{5mm}
\end{center}
\caption{Left: ROC curve for the models trained on outdoor data and tested on indoor data with descriptors taken at $35 \leq \Delta t \leq 60$. Right: ROC curve for synthetic trained models. Testing performed on indoor real descriptors.}
\label{fig:roc_out_in_synth}
\end{figure*}

\begin{figure*}[ht]
\begin{center}
\includegraphics[width=0.48\linewidth]{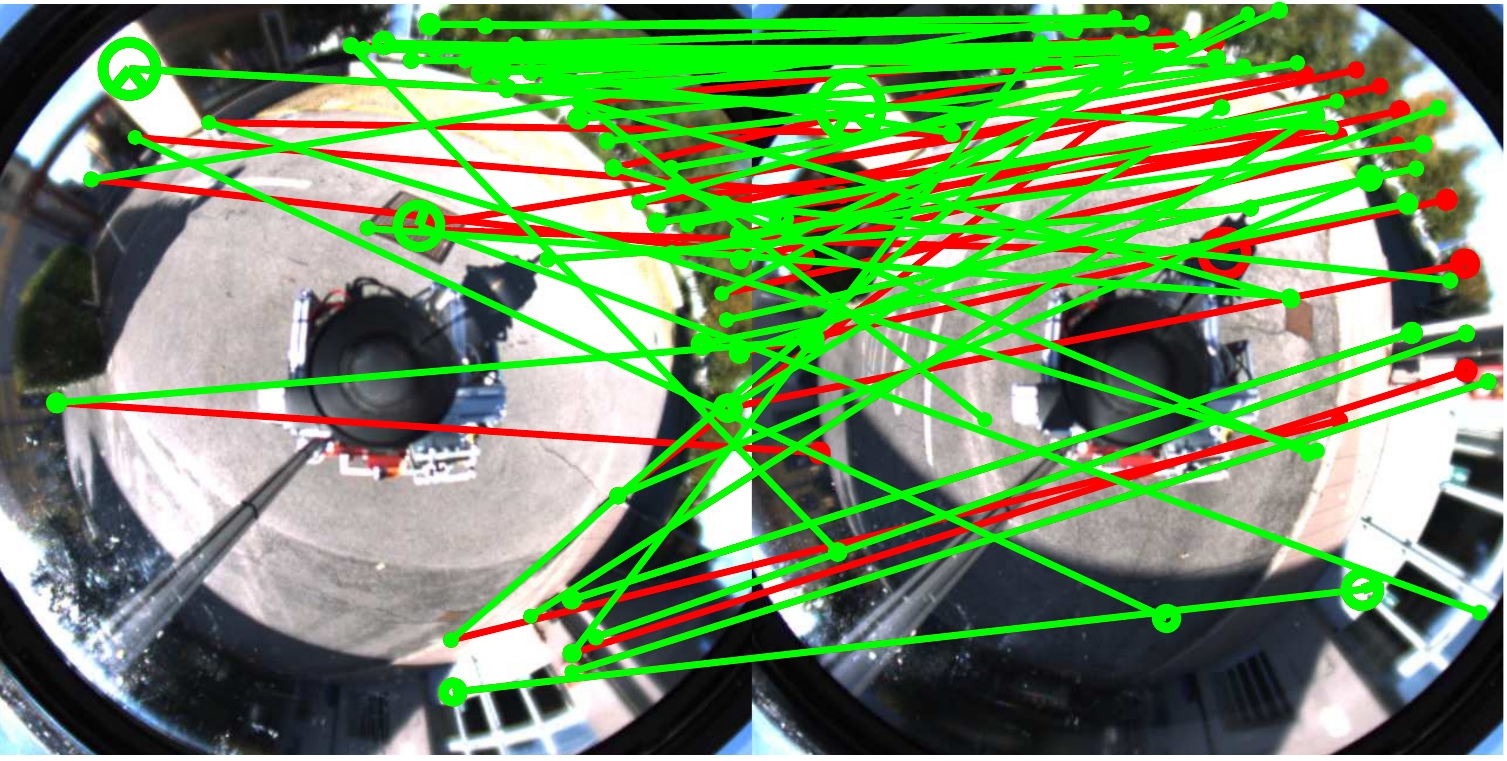} \hspace{2mm}
\includegraphics[width=0.48\linewidth]{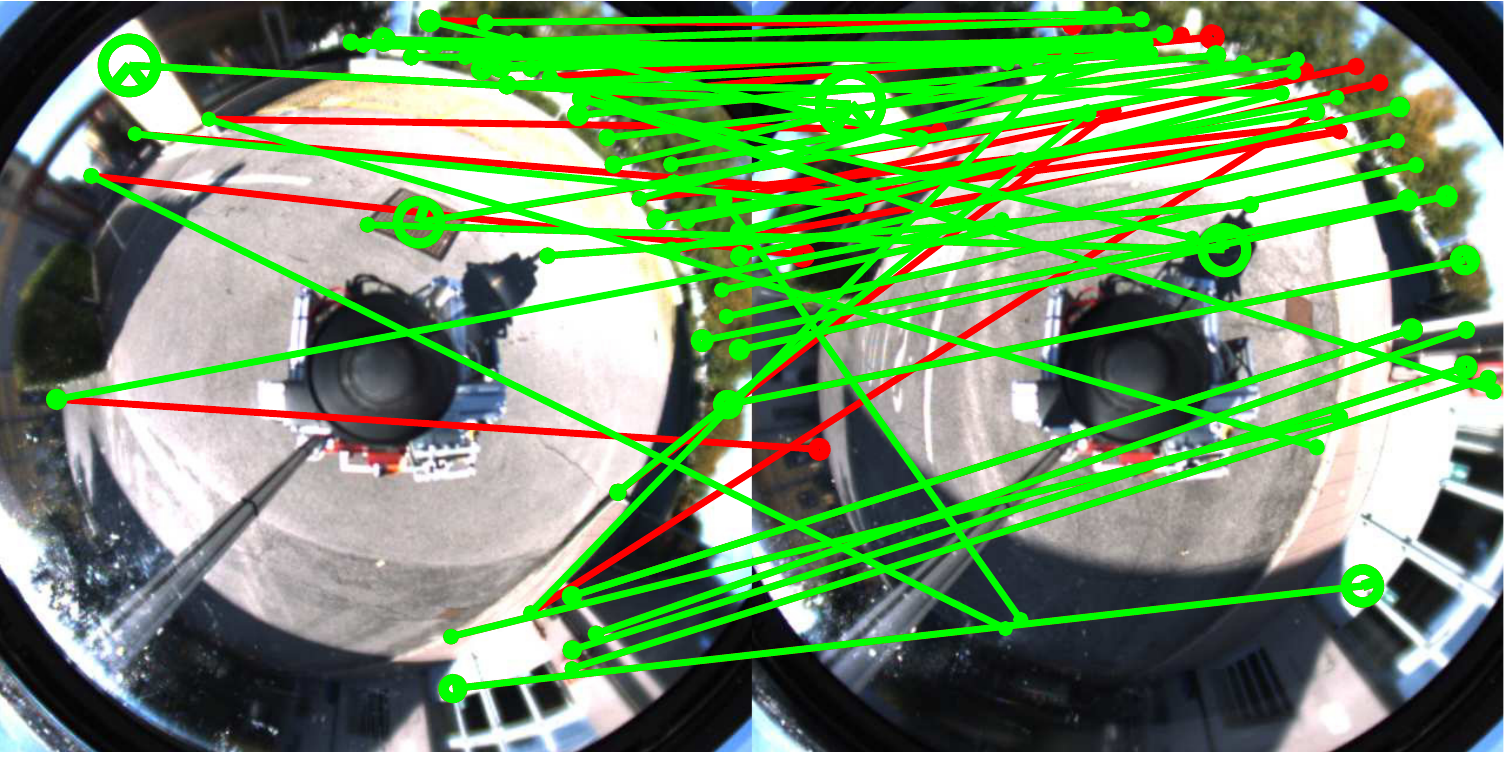}\\
\hspace{-3mm}SIFT \hspace{77mm} NNhash\\
\vspace{5mm}
\includegraphics[width=0.48\linewidth]{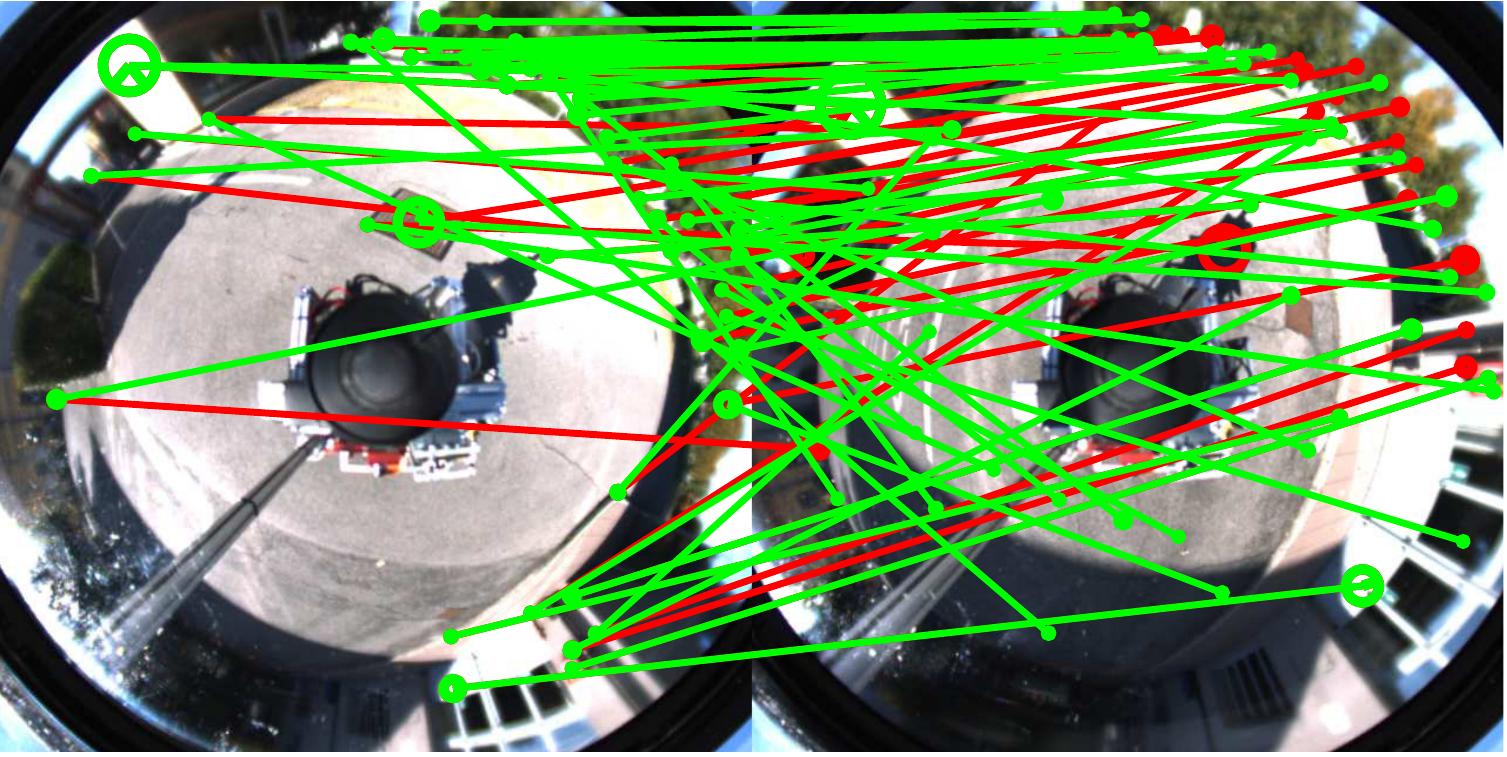} \hspace{2mm}
\includegraphics[width=0.48\linewidth]{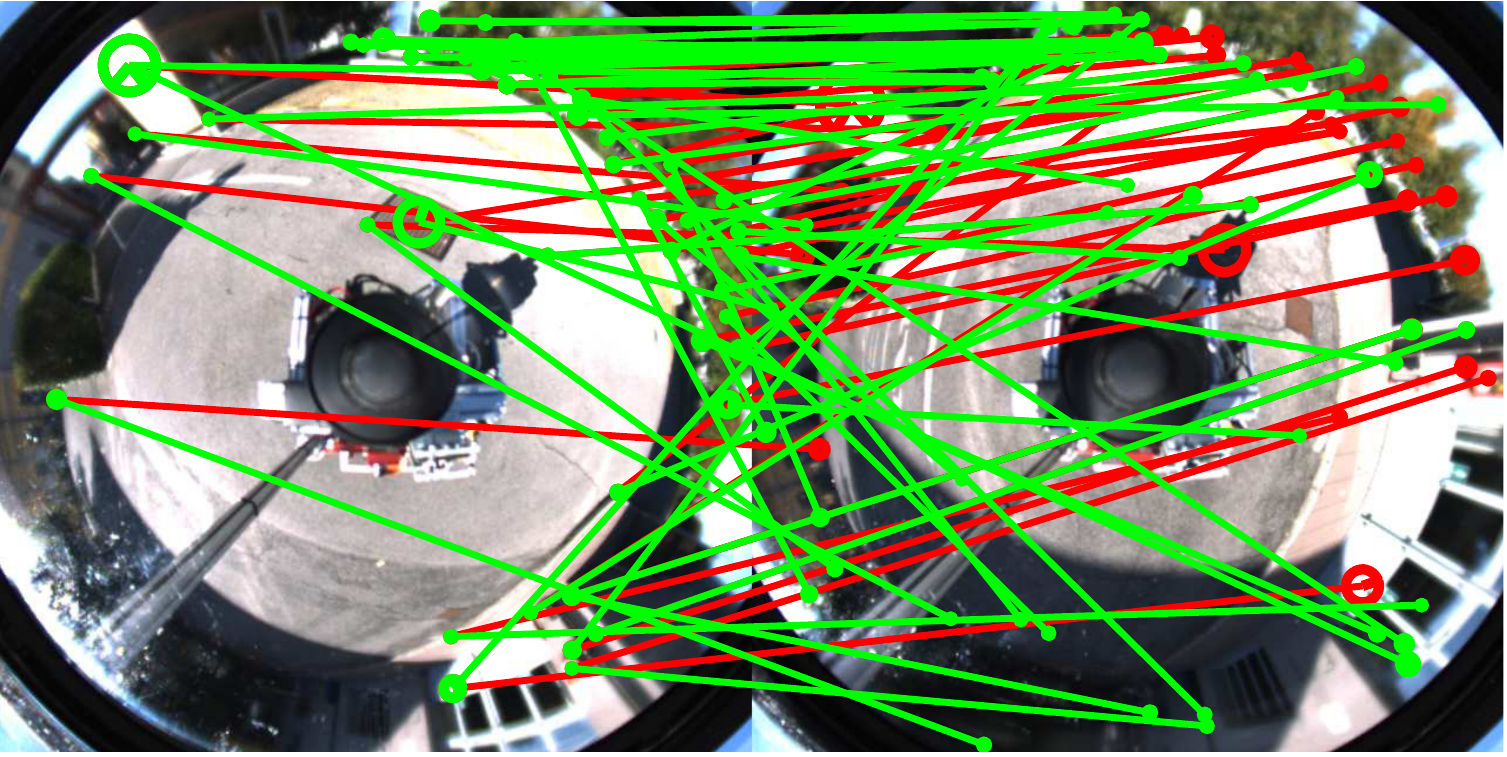}\\
\hspace{-7mm}DiffHash \hspace{77mm} SSH\vspace{-4mm}\\
\end{center}
\caption{Visual comparison of the matches produced on outdoor data with $\Delta t = 70$. Ground truth matches are plotted in red and descriptor matches (1-closest) in green. Ideally (if matching completely coincides with the groundtruth), only green lines should be visible. Interesting matches appear on the bottom-left portion of the image where NNhash learns invariance to high distortions.}
\label{fig:matches_cmp}
\end{figure*}

Figure~\ref{fig:roc_out_in_synth}-left shows the performance of descriptors trained on outdoor and tested on indoor data. We can see that even though the data used for training is very different from the one used for testing (i.e. see Figure~\ref{fig:educativeImage} and Figure~\ref{fig:matches_cmp} for a visual comparison) we achieve better performance than SIFT with just 64 bits. 
Figure~\ref{fig:roc_out_in_synth}-right shows the performance of descriptors trained on synthetic and tested on indoor data.
All learning methods perform better than SIFT. The discrepancy between NNhash and the other algorithms is less pronounced that in the real case.  

%
%
%
%
%
%
%


\section{Discussion, Conclusions, and Future Work}

We presented a new approach for feature matching in omnidirectional images based on similarity-sensitive hashing and inspired by the recent work \cite{ldahash}. We learn a mapping from the descriptor space to the space of binary vectors that preserves the similarity of descriptors on a training set. By carefully constructing the training set, we account for descriptor variability, e.g. due to optical distortions. 
The resulting descriptors are compact and are compared using the Hamming metric, offering significant computational advantage over other traditional metrics such as $L_2$. Though tested with SIFT descriptors, our approach is generic and can be applied to any feature descriptor. 

We compared several existing similarity-preserving hashing methods, as well as our NNhash method based on a neural network. 
Experimental results show that NNhash outperforms other approaches. 
An explanation to this behavior is the fact that of today's state-of-the-art similarity-preserving hashing algorithms like SSH or LDAHash solve a simplified optimization problem, whose solution does not necessarily coincide with the solution of the ``true'' non-linear non-convex problem. We showed that using a neural network, we can solve the ``true'' problem and yield better performance. 

Finally, our discussion in this paper was limited to simple embeddings of the form $\mathrm{sign}(\mathbf{P} \xx + \ttt)$ which in some cases are too simple. The neural network framework seems to us a very natural way to consider more generic embeddings using multi-layer network architectures.

\section*{Acknowledgement}

M.~B. is partially supported by the Swiss High Performance and High Productivity Computing (HP2C) grant. J.~M. is supported by Arcelor Mittal Maizi\`eres Research SA.
 D.~M. is partially supported by the EU projects FP7-ICT-IP-231722 (IM-CLeVeR).
 
{
\bibliographystyle{ieee}
\bibliography{string,vision,new,misc,omni}
}

\end{document}